\documentclass{article}
\usepackage{ijcai19}

\usepackage{times}
\usepackage{soul}
\usepackage{url}
\usepackage[utf8]{inputenc}
\usepackage[small]{caption}
\usepackage{graphicx}
\usepackage{amsmath}
\usepackage{booktabs}
\usepackage{algorithm}
\usepackage{algorithmic}
\title{Learning and Exploiting Multiple Subgoals for Fast Exploration in Hierarchical Reinforcement Learning}

\author{
    Libo Xing
    \affiliations
    Department of Computer Science, Bar-Ilan University, Israel \emails
    pcchair@ijcai19.org
}

\author{
Libo Xing$^1$
\affiliations
$^1$South China University of Technology \\
\emails
lyeby0904@gmail.com
}

\begin{document}

\maketitle

\begin{abstract} 
  Hierarchical Reinforcement Learning (HRL) exploits temporally extended actions, or options, to make decisions from a higher-dimensional perspective to alleviate the sparse reward problem, one of the most challenging problems in reinforcement learning. The majority of existing HRL algorithms require either significant manual design with respect to the specific environment or enormous exploration to automatically learn options from data. To achieve fast exploration without using manual design, we devise a multi-goal HRL algorithm, consisting of a high-level policy Manager and a low-level policy Worker. The Manager provides the Worker multiple subgoals at each time step. Each subgoal corresponds to an option to control the environment. Although the agent may show some confusion at the beginning of training since it is guided by three diverse subgoals, the agent’s behavior policy will quickly learn how to respond to multiple subgoals from the high-level controller on different occasions. By exploiting multiple subgoals, the exploration efficiency is significantly improved. We conduct experiments in Atari’s \cite{bellemare2013arcade} Montezuma’s Revenge environment, a well-known sparse reward environment, and in doing so achieve the same performance as state-of-the-art HRL methods with substantially reduced training time cost.

\end{abstract}

\begin{figure*}[ht]
	\centering
	\includegraphics[width=0.9\linewidth]{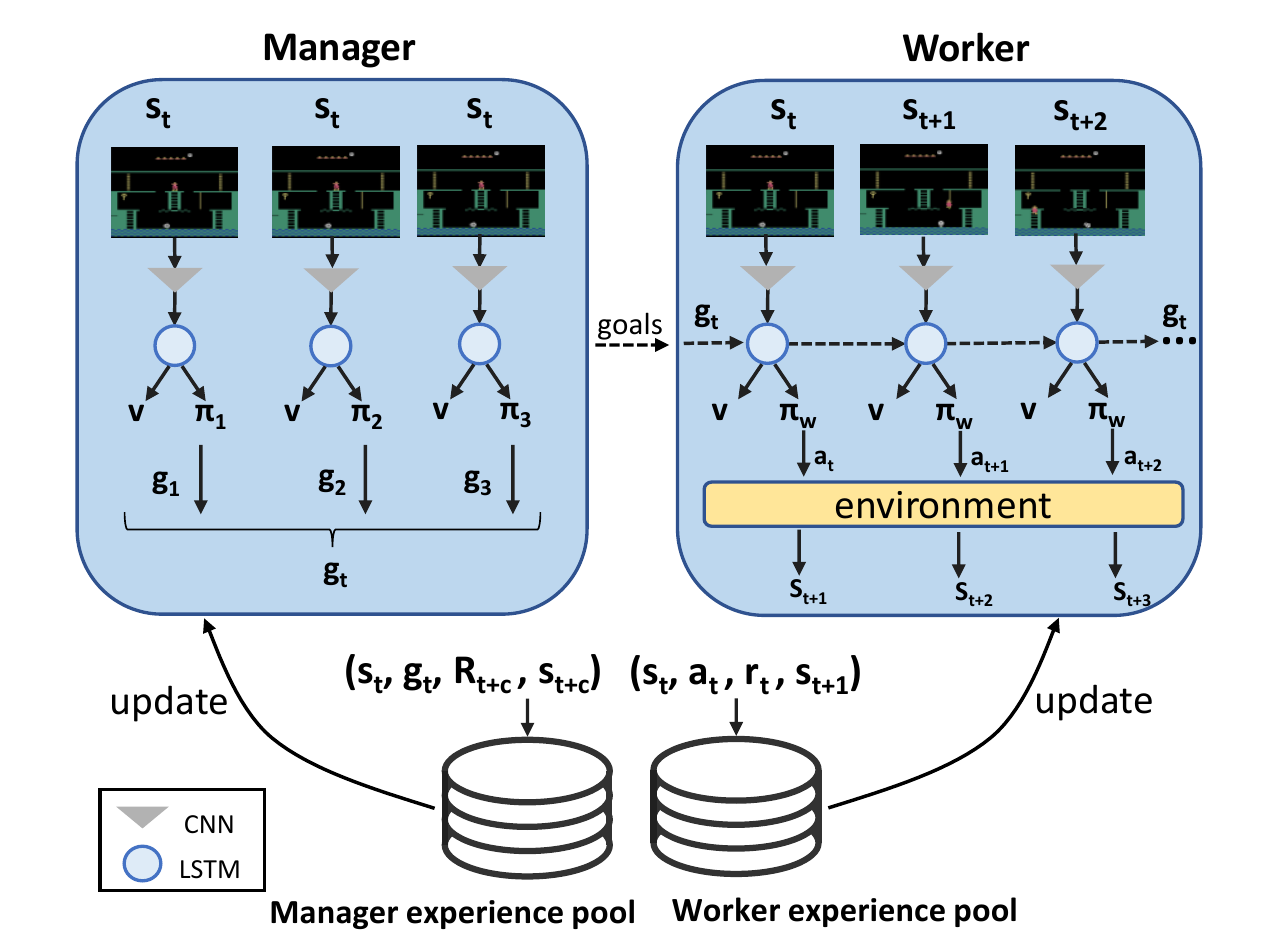}
	\caption{The schematic illustration of our MGHL.}
	\label{fig:false-color}
\end{figure*} 

\section{Introduction}
Deep Reinforcement Learning (DRL) \cite{mnih2015human} has significantly contributed to several areas of artificial intelligence. AlphaZero \cite{silver2017mastering} combined DRL with tree search to achieve superhuman performance in the extremely challenging strategic game, Go. Recently, AlphaStar \cite{DeepMind}, which is also based on DRL, defeated the top human players of StarCraft. DRL has, therefore, been integral to breakthroughs in high complexity and tactical problem-solving. It has also become the mainstream algorithm in reinforcement learning and remains a focus for further development.

The problem of sparse rewards has always been one of the hardest challenges in reinforcement learning including DRL. In sparse reward environments, it is very difficult for an agent to find the reward through random attempts at completing a task. This means that the agent needs to perform many actions sequentially before obtaining the reward, making it difficult for the agent to learn the necessary actions. Long-term credit assignment remains elusive for DRL algorithms in sparse reward environments.

Hierarchical Reinforcement Learning (HRL) introduces a multi-level structure to reinforcement learning, and aims  to train on successive and higher temporal abstractions. The main problem that the hierarchical structure needs to solve is how to effectively learn temporal abstractions. Prior studies have aimed to find effective expressions of subgoals and then choose the right subgoal as the high-level action at each time step to instruct the behavior policy. However, defining subgoals is a complicated problem. Two popular methods have been used to define subgoals: (i) by using prior knowledge of the specific environment to find salient events as subgoals; however, this method often requires intensive manual design so it lacks generalizability; and (ii) by creating abstractions autonomously from data as subgoals or designing subgoals from a primitive and general perspective to reduce the manual design. These methods suffer from the challenge of exploring efficiency.

In this paper, we choose to automatically generate subgoals from data to improve generalizability. To solve the problem of exploration efficiency inherent in previous methods, we propose a multi-goal HRL algorithm. Our algorithm is inspired by the feature-control agent \cite{dilokthanakul2017feature} and Multi-DDPG agent \cite{yang2017multi}. 

With respect to the feature-control agent, we find that designing auxiliary control tasks into subgoals can speed up the learning of the HRL agent. In many cases, the positive effects of different auxiliary control tasks on the agent do not conflict. Therefore, we consider exploiting multiple auxiliary tasks at the same time. 

With respect to the Multi-DDPG agent, we find that training multiple goal-policies simultaneously through the same training process can significantly improve training efficiency. This demonstrates the feasibility of simultaneously exploiting multiple auxiliary tasks. Hence, we design a scheme that takes advantage of all the subgoals representing the auxiliary control tasks to speed up exploration. Our algorithm adopts the idea \cite{jaderberg2016reinforcement} that the auxiliary control tasks can provide the agent with useful features to manipulate the environment.

Our main contribution is that we devise a new HRL algorithm that takes the advantage of using multiple subgoals simultaneously. (1) By defining subgoals as an auxiliary control task and simultaneously using multiple subgoals to instruct our behavior policy, our algorithm improves the exploration ability of the agent. (2) By designing generalized subgoals that require little prior knowledge of the environment, we meliorate the generalizability of the algorithm. We have tested our algorithm in Atari’s multiple environments, especially Montezuma’s Revenge, the most complex sparse reward environment. Compared with other public HRL algorithms, we show that the new algorithm substantially benefits from exploring efficiency.

\begin{figure*}[htbp]
	\begin{center}
		\includegraphics[width=0.8\linewidth]{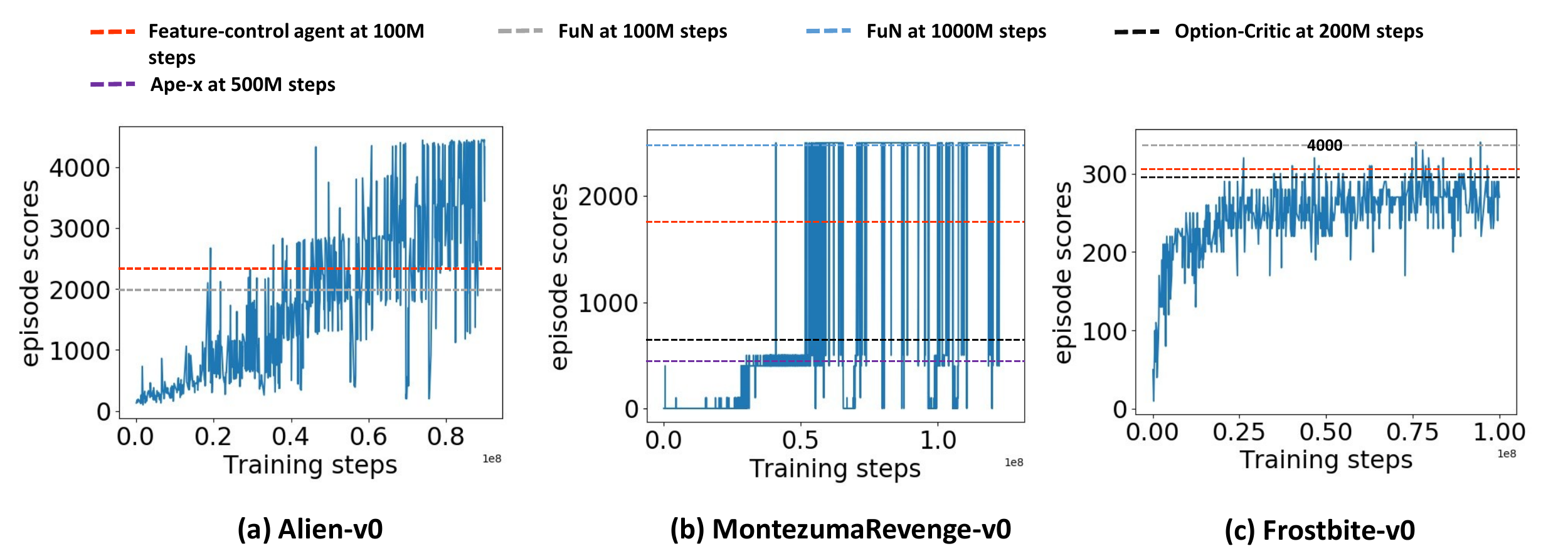}
		\caption{Comparision to state-of-the-art systems. \textbf{a)} Learning curve on intensive-reward environment Alien. \textbf{b)} Learning curve on sparse-reward environment Montezuma's revenge. \textbf{c)} Learning curve on changing environment Frostbite. The value is the average per episode score of top 5 agents. We use all the three subgoals.}
	\end{center}
\end{figure*}

\section{Related work}
Designing a simple and efficient HRL algorithm is a longstanding problem in reinforcement learning. Many works have considered improving the exploration efficiency from different perspectives, as summarized below.

\subsection{All-goals updating}
It is intuitive to train agents with different goals to improve efficiency. "All-goals updating" \cite{kaelbling1993learning} was originally proposed by Kaelbling in 1993 and has since been further developed several times. All-goals updating proposes a method of simultaneously updating many goal-conditional action-value functions using a single stream of experience. Many previous algorithms have investigated how to define goals and how to train them based on all-goals updating. Representatively, Vivek et al. \cite{veeriah2018many} designed a method called “many-goals”, in which the agent generated many goals in the form of raw pixel observations and updated the value function of several goals during the training process. However, these algorithms calculate the value function of each goal just to select the best goal and do not consider taking advantage of many goals by exploiting these goals simultaneously to improve exploration efficiency. In our algorithm, we consider training the agent with multiple goals at the same time in HRL to improve the exploration efficiency.

\subsection{Intrinsic motivation}
Intrinsic motivation aims to provide qualitative guidance for exploration. In HRL algorithms, subgoal is the manifestation of intrinsic motivation. How to define subgoals greatly affects the agent’s exploration efficiency. Embodying an agent with a form of intrinsic motivation has been explored in several previous works. The FuN \cite{vezhnevets2017feudal} algorithm is a well-known two-layer hierarchical reinforcement learning algorithm that defines the subgoals as the direction in the latent state space. The HDQN \cite{kulkarni2016hierarchical} algorithm defines some target locations as subgoals by manual design. These are different manifestations of intrinsic motivation. However, how to introduce multiple subgoals to these algorithms is unclear. In some works, intrinsic motivation is defined to find bottleneck states \cite{csimcsek2004using,mcgovern2001automatic,nachum2018data} in the environment. However, discovering the bottleneck states requires a lot of environmental statistics, so it can be impossible in complex environments, especially in sparse reward environments.

\subsection{Auxiliary control tasks}
The UNREAL \cite{jaderberg2016reinforcement} architecture proposed the concept of auxiliary control tasks and designed pixel-control and feature-control tasks in the vision domain. These two auxiliary tasks greatly improved the performance in the Atari environment. The feature-control agent designed these two auxiliary control tasks in the form of intrinsic motivation; the subgoals of the algorithm were correspondingly executed to change the pixels of the specified area or the specified higher-order environmental features. Based on the idea of feature-control, we have designed a new auxiliary subgoal, direction-control. Compared to the auxiliary control tasks mentioned above, the new auxiliary control task is more concise and closer to the agent itself.

Inspired by HDQN and FuN, we use a two-layer structure that includes the Manager, which is used to create subgoals, and the Worker, which is used to behave directly. What makes our algorithm unique is the way in which multiple auxiliary tasks are applied simultaneously, including the two auxiliary tasks introduced in UNREAL and the newly designed direction-control task. Our algorithm substantially improves the exploring efficiency of the agent.

\begin{figure*}[ht]
	
	\centerline{\includegraphics[width=0.8\linewidth]{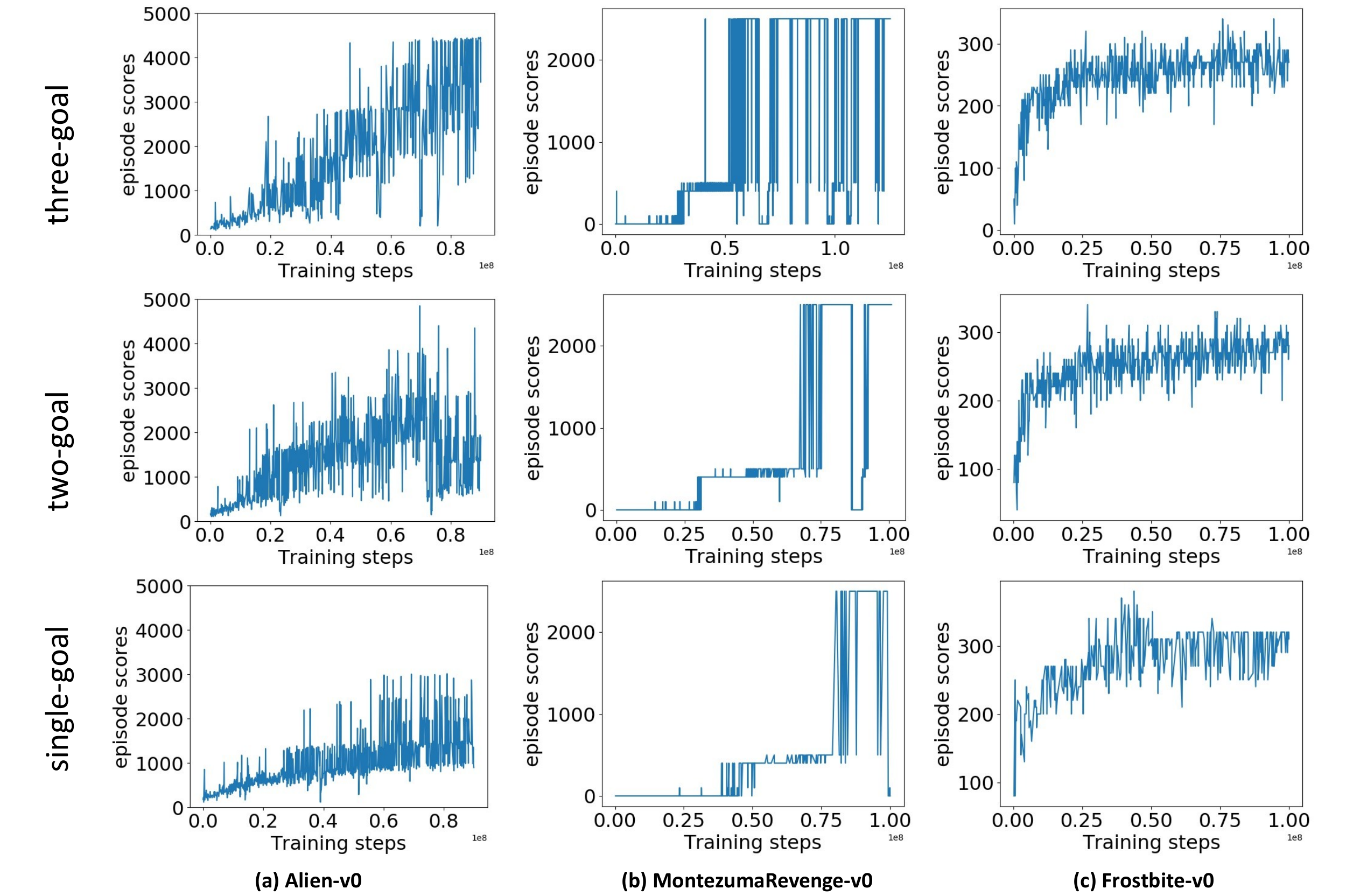}}
	\caption{Comparison of various numbers of subgoals on Alien, Montezuma's revenge and Frostbite.}
\end{figure*}

\section{The model}
This section details the proposed Multi-Goal HRL (MGHL) algorithm to handle the problem of sparse rewards. We make use of several useful subgoals at the same time and no longer choose between them. Our algorithm consists of two parts: a high-level policy Manager and a low-level policy Worker. Each subgoal-generator in the Manager module is a separate actor-critic structure, which we call the goal-policy. Each goal-policy contains a goal-actor and a goal-critic. The Worker receives multiple subgoals from the Manager at the same time and then performs actions. As long as the action of the Worker meets the instruction of any one of the subgoals, the corresponding intrinsic reward will be obtained. For multiple subgoals, the intrinsic rewards obtained by the Worker can be superimposed. Although different subgoals may be contradictory, the Worker does not need to satisfy all of them to get rewards. Through continuous training, the agent will become more and more awake. Figure 1 provides an overview of the algorithm.

\subsection{Reinforcement Learning setting}
We adopt the standard reinforcement learning setup, where an agent interacts with the environment in discrete time steps. At each time step, the agent receives an observation ${s_t}$ provided by environment and performs an action according to behavior policy $\pi :S -> A$. Then we will get the next state ${s_{t + 1}}$ and the reward ${r_t}$.In standard reinforcement learning setup, the agent’s goal is to maximize the expected future return ${G_t} = \sum\limits_{k = 0}^\infty  {{\lambda ^k}{R_{t + k + 1}}} $, where $\lambda  \in (0,1)$ is the discount factor and ${t}$ is the current time step. The state-action value function ${Q^\pi }(s,a) = {E_{s'}}[R + \lambda {Q^\pi }(s',a')|s,a]$ is the expected return if the agent perform action ${a}$ at the state ${s}$ and the perform action following the policy $\pi$.
Value-based reinforcement learning algorithms commonly approximate the state-action value function using parameters $\theta$ and update $\theta$ to minimize the Temporal-Difference error according to Bellman  equation, ${L_Q}({s_t},{a_t}) = {({Q^\pi }({s_t},{a_t}) - r - {Q^\pi }({s_{t + 1}},{a_{t + 1}}))^2}$. Unlike value-based algorithms, the algorithms with actor-critic architecture combine a behavior policy and state-action action within one framework so that the agent can solve more complex tasks by learning complex policies. The Asynchronous Adavantage Actor-Critic algorithm maintains a policy $\pi ({a_t}|{s_t};\theta )$ and an estimate of the value function $V({s_t},{\theta _v})$, both the policy and value function use the n-step returns in the forward view. The policy objectiove function’s gradient including the entropy regularization term takes the form :
\begin{equation}
	{\nabla _{\theta '}}\log \pi ({a_t}|{s_t};\theta ')({R_t} - V({s_t};{\theta _v})) + \beta {\nabla _{\theta '}}H(\pi ({s_t};\theta ')),
\end{equation}
where ${H}$ is the entropy regularization term. In asynchronous advantage actor-critic (A3C) \cite{mnih2016asynchronous} algorithm, many parallel agents interact with many instances of the environment, only accumulating the gradient of each instance to update the global agent. In this way, A3C speeds up training and achieves better results.

\subsection{Goal setting}
Here we provide the details of the subgoals used by our algorithm. Each subgoal corresponds to an auxiliary control task.

\subsubsection{Pixel control}  
In video games, the obtainable environmental information is the current image frame. A large change in the pixel of an image block usually means that an important event has occurred in this block, so that pixel block could be used as a subgoal setting. We split the input image frame into multiple blocks and choose one block as a subgoal at each time. Once the agent’s action causes the pixels of the image to change, the agent gets rewarded. The feature-control agent proposed a way to define intrinsic reward for pixel control: the pre-processed 84*84 input image was divided into 36 pixel blocks of size 14*14. The Worker needs to perform an action to change the pixels in the pixel block specified by the pixel-controlled policy of the Manager so that it can be rewarded. The subgoal of our pixel-control policy is to tell the Worker which pixel block to change as much as possible. We use the same setting as the feature-control agent:
\begin{equation}
	{r^{{\mathop{\rm pc}} }}(k) = \eta \frac{{||{h_k} \odot ({s_t} - {s_{t - 1}})|{|^2}}}{{||{s_t} - {s_{t - 1}}|{|^2}}}.
\end{equation}

\subsubsection{Direction control}
Agents always explore by moving around in the environment. We can set a direction-related subgoal for the Worker, telling the agent which direction to explore. Direction control is the most basic guidance for agents. In video games, we define five directions, north, south, east, west, and $^\circ$, where $^\circ$ represents standing still. At each time step $t$, if the agent moves in the direction specified by the Manager, the intrinsic reward is set to 0.01, otherwise it is set to 0:
\begin{equation}
	{r^{{\mathop{\rm dc}} }}(k) = 1({a_t} \in k)*0.01.
\end{equation}

The direction-related subgoal provides the agent with a simpler way to obtain a reward. This makes the agent more flexible in the training process so that the agent can better explore the environment.

\subsubsection{High-order feature control}  
Convolutional neural networks can obtain high-level features of images by convolving the input data. We can think of each convolutional network layer as a high-level representation of the image, and each feature map corresponds to a higher-order pixel value. Similar to pixel control, we use a subgoal to have more abstract control over the environment. The feature-control agent proposed one way to measure how much a feature can be controlled through intrinsic reward, which can be written as:
\begin{equation} 
	{r^{{\mathop{\rm fc}} }}(k) = \eta \frac{{||{f_k}({s_t}) - {f_k}({s_{t - 1}})||}}{{\sum\nolimits_{k'} {||{f_{k'}}({s_t}) - {f_{k'}}({s_{t - 1}})||} }}.
\end{equation}

\subsection{Training}
Based on the proposed A3C algorithm and feature-control agent, we extend the algorithm from single-goal to multi-goal learning. Our model consists of two parts: the Manager module and the Worker module. The Worker module has one actor that receives multiple subgoals from the Manager module and interacts directly with the environment based on the subgoals given by the Manager. The Manager module trains multiple goal-actors, each corresponding to a subgoal and with an independent critic. These actors are trained at the same time and are learned from the same experience. The Manager is trained to get more extrinsic rewards.

 We use the policy gradient method to train the Manager end-to-end through the actions performed by the Worker and the rewards given by the environment, since our agent is fully differentiable. All goal-actors in the Manager module are updated independently. We formalize the update rule for each goal-actor ${\pi ^{{M_i}}}$ as:

\begin{equation}
	\nabla {g_i}(t) = A_t^{{M_i}}{\nabla _\theta }\log {\pi ^{{M_i}({x_t})}}
\end{equation}

\begin{equation}
	A_t^{{M_i}} = R_t^{ext} - V_t^{{M_i}}({x_t};\theta ),
\end{equation}
where ${i}$ is identity number of the goal-actor and $R_t^{ext}$ is the extrinsic discounted return. The extrinsic discounted return can be written as:
\begin{equation}
	R_t^{ext} = \sum\nolimits_{k = 0}^\infty  {{\gamma ^k}r_{t + k + 1}^{ext}}.
\end{equation}

Since the Worker treats each subgoal equally, all the goal-actors are trained in the same way through the actual sampling experience of the Worker. We do not distinguish subgoals produced by different goal-actors and we do not care about which subgoal is responsible for the behavior of the Worker. The Worker is a standard actor-critic architecture, where critic and actor are trained synchronously. All subgoals are concatenated together and merged with the environment information as input to the Worker. Unlike the Manager’s goal-actors, the Worker uses a mixed reward, which consists of an extrinsic reward and an intrinsic reward, as the supervising signal for training. 
The intrinsic reward is defined as:
\begin{equation}
	r_t^{{\mathop{\rm int}} } = r_t^{pc} + r_t^{fc} + r_t^{dc}.
\end{equation}

We also used an extrinsic reward as an auxiliary reward to achieve better performance. The final reward of Worker can be formalized as:
\begin{equation}
	{r_t} = \frac{{(1 - \alpha )}}{2}r_t^{{\mathop{\rm int}} } + \alpha r_t^{ext},
\end{equation}
where ${\alpha}$ is the parameter to adjust the ratio between intrinsic reward and extrinsic reward. Note that we use a coefficient of $\frac{{1 - \alpha }}{2}$ for different numbers of subgoals without changing the coefficient with the number of subgoals.

\subsection{Model Details}
We next provide more details of the model described above. Both in the Manager and Worker, observations are first fed into an encoder module, which is a convolutional neural network (CNN) followed by a fully connected layer. The CNN contains two convolutional layers: a first layer with 16 8*8 filters of stride 4 and a following layer with 32 4*4 filters of stride 2. Each layer is followed by a ReLU non-linearity layer. In the Manager, the output of the fully connected layer merges with the previous reward, and the previous action is then fed into an LSTM layer. All the subgoals produced by the Manager module are converted to one-hot vector form, which specifies the way in which the Worker gets intrinsic rewards. These are concatenated with the output of the Worker’s fully connected layer and then fed into an LSTM layer. All the LSTM layers have 256 hidden units. We use the A3C method for all experiments. For our agent, we use 32 threads for the A3C training method.

\begin{figure*}[htbp]
	\centering
	\includegraphics[width=\linewidth]{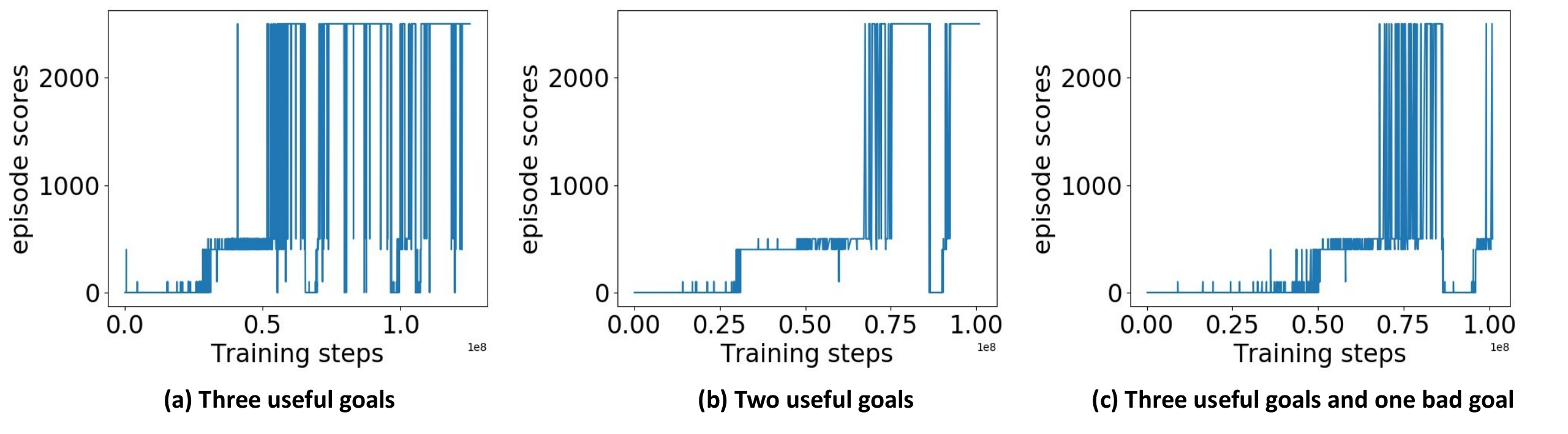}
	\caption{The performance of adding a random subgoal in the Manager. This random subgoal is kept random throughout the training, and the agent can get a fair reward as long as the action matches this random subgoal.}
\end{figure*}

\section{Experiments and Results}

\subsection{Atari and Montezuma’s Revenge}
The Arcade Learning Environment (ALE) has had a big impact on reinforcement learning research. Here we used OpenAI Gym’s Atari game environment, which wraps the ALE with some necessary adjustment for our algorithm. Montezuma’s Revenge is one of the most difficult game environments in ALE, with sparse rewards and complex terrain. In the first scene of Montezuma’s Revenge, the agent needs to travel a long distance to get the key to get a small reward, before finding and opening the door to get more rewards. The sparseness of rewards means that agents face a large exploration space. 
The Atari Montezuma’s Revenge environment provides an RGB image of the screen as the observation, which is an array of shape (210,160,3). Our model only uses the observation as input without any additional information. We tested our algorithms on multiple game environments in Atari. Uniformly, we used BPTT=20 for the Manager and BPTT=100 for the Worker without adjustments for specific environments. In order to reflect differences in the extrinsic rewards, we divided the actual extrinsic reward by 100 as our final extrinsic reward instead of clipping it to [-1,1], which is applied by many algorithms. We set the discount factor $\gamma $ to 0.99 for both the Manager and Worker.

\subsection{Comparison to state-of-the-art hierarchical RL systems}
In this experiment, we compared our agent with the state-of-the-art HRL algorithm in Alien, Montezuma’s Revenge, and Frostbite, which separately represent sparse rewards, intensive rewards, and scene-changing environments. The three algorithms compared here are FuN, feature-control, and the option-critic algorithm. In order to illustrate the effectiveness of the hierarchical reinforcement learning algorithm in the sparse reward environments, we also compare our agent with the non-hierarchical state-of-the-art RL algorithm Ape-X \cite{horgan2018distributed} in the Montezuma’s Revenge environment. In fact, some reinforcement learning algorithms have achieved excellent results in sparse environments, such as RLED \cite{pohlen2018observe}. However, since they used demonstration data or too much manual design and our hierarchical reinforcement learning algorithm is designed to minimize manual design, a fair comparison is difficult.

Our results are shown in Figure 2, where we evaluate agents with full subgoals in each environment. In Alien our agent scored nearly twice the score of FuN and feature-control agent under the same number of training steps. As can be seen from (b), in the Montezuma's revenge environment, the advantage of our algorithm is mainly the efficiency of exploring. Our agent uses fewer training steps reaching the same maximum scores than other agents. This also proves that our multi-goal framework greatly enhances the exploration performance. In the Frostbite environment, our algorithm achieves similar performance to Option-critic and feature-control agent, but worse than FuN. 

In addition, in the non-sparse reward environment of Alien and Frostbite, the performance of our algorithm and other hierarchical reinforcement learning algorithms is far worse than Ape-x. This is a limitation of the hierarchical reinforcement learning algorithm.

\subsection{Ablation study}
We next evaluated our agent using different numbers of subgoals, the aim being to verify whether our multi-goal setting is necessary. We tried reducing the subgoals generated by the Manager module from three to one for the Worker module. The settings were: (i) pixel-control subgoal; (ii) pixel-control and feature-control subgoals; and (iii) pixel-control, feature-control, and direction-control subgoals. The single-goal agent actually evolved into the feature-control agent. The actor-critic network for each subgoal of the Manager is independent and does not share any parameters. We experimented with many different configurations for the Worker’s reward combination, mainly the combination of multiple intrinsic rewards. The results shown are the performances of the best reward combinations.

Figure 3 shows the results of each setting on some Atari games. In the Montezuma’s Revenge environment, our agents eventually reached the same score, but the number of steps required for training greatly increased as the number of subgoals decreased. In Alien, our three-goal agent finally achieved twice the performance of the single-goal agent in the same number of training steps. In Frostbite, all the three settings delivered similar performance.

These results show that our multi-goal setting is necessary. However, the form of the subgoals can be more general, since our pixel-control subgoal is not applicable in many non-image input agents.

\subsection{Explore the robustness of our framework}
In our multi-goal algorithm, if the lower bound of the algorithm depends on the worst subgoal, then our algorithm will be difficult to apply. To illustrate the robustness of our algorithm, we added a bad subgoal to the algorithm, which is a random instruction. Our results are shown in Figure 4, and we note the following: although the performance is reduced after adding a bad subgoal, the performance of the three-goal agent is still satisfactory. This shows that a useless subgoal of the Manager module does not cause catastrophic damage to the performance of the entire algorithm. This also reflects the robustness of our algorithm.

\section{Conclusion}

Here we propose a Multi-Goal HRL algorithm with the aim of improving the efficiency of exploration, especially in sparse reward environments. In our algorithm, we apply multiple subgoals to our behavior policy and no longer choose from them. These subgoals are learned from data directly and thus do not require manual design. Our experiments show that our algorithm outperforms existing HRL algorithms in Montezuma’s Revenge, with particular sparse rewards and some environments with intensive rewards. Although our results are not perfect, they prove the feasibility of this algorithm which provides a new line for the study of HRL. Our future work will focus on how to generalize subgoals and how to learn the impact of each subgoal on Workers at different times.

\bibliographystyle{named}
\newpage
\bibliography{reference}

\end{document}